\documentclass[letterpaper, 10 pt, conference]{ieeeconf}

\usepackage{graphics} 
\usepackage{epsfig} 
\usepackage{times}
\usepackage{amsmath} 
\usepackage{amssymb}  
\usepackage{verbatim}

\usepackage{amsthm}
\usepackage{bbm}
\usepackage{amsfonts,balance}
\usepackage{bm}
\usepackage{graphicx}
\usepackage{caption}
\usepackage{subcaption}
\usepackage{algorithm}
\usepackage{algorithmic}
\usepackage{color}
\usepackage{setspace}

\IEEEoverridecommandlockouts

\usepackage[svgnames]{xcolor} \definecolor{DarkGreen}{rgb}{0,0.5,0}
\definecolor{DarkRed}{rgb}{0.75,0,0}

\makeatletter \let\NAT@parse\undefined \makeatother
\usepackage[sort,compress]{cite}

\title{\LARGE \bf
Active Perception in Adversarial Scenarios using Maximum Entropy\\ Deep Reinforcement Learning} 

\author{Macheng Shen$^{1}$, Jonathan P.\ How$^{2}$
\thanks{$^{1}$M.~Shen is with the Department of Mechanical Engineering, 
        Massachusetts Institute of Technology (MIT), 77 Massachusetts Ave., Cambridge, MA, USA.
        {\tt macshen@mit.edu}}%
\thanks{$^{2}$J.~How is with the MIT Department of Aeronautics and Astronautics, 77 Massachusetts Ave., Cambridge, MA, USA.
        {\tt jhow@mit.edu}%
}}

\begin{document}

\maketitle
\thispagestyle{empty}
\pagestyle{empty}
\begin{abstract}
We pose an active perception problem where an autonomous agent actively interacts with a second agent with potentially adversarial behaviors. Given the uncertainty in the intent of the other agent, the objective is to collect further evidence to help discriminate potential threats. The main technical challenges are the partial observability of the agent intent, the adversary modeling, and the corresponding uncertainty modeling. Note that an adversary agent may act to mislead the autonomous agent by using a deceptive strategy that is learned from past experiences. We propose an approach that combines belief space planning, generative adversary modeling, and maximum entropy reinforcement learning to obtain a stochastic belief space policy. 
By accounting for various adversarial behaviors in the simulation framework and minimizing the predictability of the autonomous agent's action, the resulting policy is more robust to unmodeled adversarial strategies. This improved robustness is empirically shown against an adversary that adapts to and exploits the autonomous agent's policy when compared with a standard Chance-Constraint Partially Observable Markov Decision Process robust approach.
\end{abstract}

\section{Introduction}
We are interested in an active perception problem, where an autonomous agent interacts with a second agent (which we refer to as the opponent in the following context), whose intention is unknown. The objective is evidence accumulation to help identify the intent of the opponent. In order to achieve this goal while ensuring safety, the autonomous agent has to reason about the possible reactions of the opponent in response to its exploring actions and maximize the information gained from this interaction. This type of problem can find applications in various domains such as urban security and humanitarian assistance.

One field of research closely related to our work is Threat Assessment (TA). One approach to TA is Adversarial Intention Recognition (AIR), i.e., recognizing the intentions of a potential adversary based on the observations of its actions and states. Early works in AIR rely on a library of adversarial plans \cite{avrahami2014keyhole}, \cite{li2012approach}, explicitly encoding a set of expected adversarial behaviors. This approach could suffer from incompleteness of the behavior library. Most recent advancements in AIR use a two-phase approach combining generative plan recognition and game theoretic planning \cite{le2015generative}, \cite{le2016adversarial}, \cite{ang2017game}. In the first phase, a probability distribution over the potential intentions of the adversary is inferred by solving an inverse planning problem given an agent action model and a set of possible intentions. In the second phase, a set of stochastic games, each corresponding to one specific hostile intention, are solved, and the Nash Equilibrium policy is obtained as the best response to the adversary. This existing framework, however, does not account for adversarial strategies that are actively against the intention recognition, e.g., deceptive behaviors.
\begin{figure}[t]
\centering
\includegraphics[width=0.9\columnwidth]{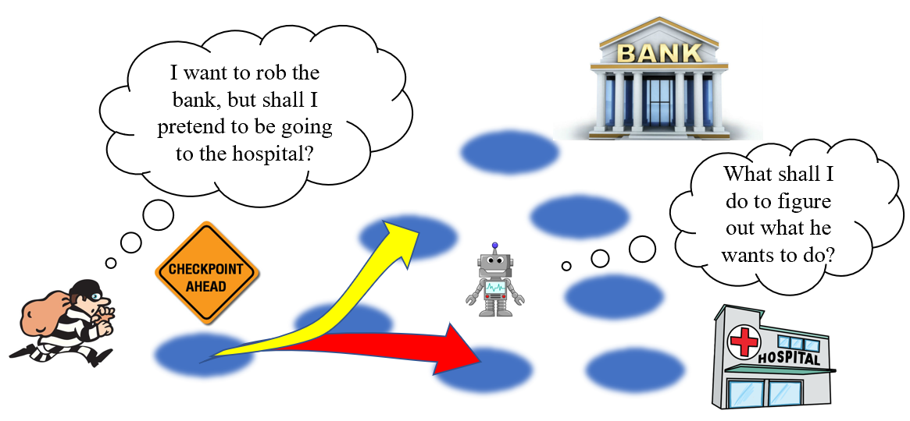} 
\caption{Illustration of the active perception challenge: The autonomous agent wants to infer the intention of the opponent, while accounting for its possible deceptive behaviors.}
\vspace{-.2in}
\label{illustration}
\end{figure}

Our active perception problem differs from the AIR problem in that: 1) the opponent is not necessarily adversarial, but could be self-interested; and 2) the primary objective is discriminating potential threat through evidence accumulation rather than defending against adversarial attacks. Despite these subtle differences, these two problems share the same challenge of reasoning about the hidden intention (partial observability) of the potential adversary subject to the modeling uncertainty of the opponent behavior, which is illustrated in Fig.~\ref{illustration}. Therefore, we expect that the techniques in our approach are also applicable to AIR problems.

 Belief space planning provides a principled framework to perform optimally in a partially observable world. Generative adversary modeling provides a variety of adversarial behaviors against which the autonomous agent can optimize its policy so that robustness is gained. This also avoids the requirements of domain experts and the difficulty of handcrafting a library of adversary behaviors. Reinforcement learning makes it possible to learn a policy from a black-box simulator that represents a complicated mixture of behaviors, which could be difficult in planning-based approaches. The maximum entropy framework minimizes the exploitability of the resulted policy by minimizing the predictability of the actions, thus making it more robust to un-modeled adversarial strategies. 

To summarize, the contribution here is the development of a scalable robust active perception method in scenarios where a potential adversary opponent could be actively hostile to the intent recognition activity, which extends and outperforms the POMDP methods.
 
\section{Related works}
We review three related fields of research:
\begin{enumerate}
\item POMDP: finds a deterministic optimal policy in a partially observable scenario given a reasonably accurate agent model;
\item Game theory: finds robust Nash Equilibrium policies given a payoff (reward) profile encoding the preference (dictated by the intent) of the agents, without the requirement of an agent model;
\item Deep reinforcement learning: enables learning an optimal policy from a simulator, the building block that our work is based on. 
\end{enumerate}   
\subsection{POMDP frameworks}
The POMDP framework provides a principled approach to behave optimally in a partially observable environment. 
One restriction of this framework is that it assumes a fixed reactive probabilistic model of the opponent, implying stationary behavior without rationality. 
To mitigate performance degradation due to modeling uncertainty, existing approaches include Bayesian-Adaptive POMDP (BA-POMDP) \cite{ross2011bayesian}, \cite{katt2018learning}, robust POMDP \cite{osogami2015robust}, Chance-constrained POMDP (CC-POMDP) \cite{santana2016rao}, and Interactive-POMDP (I-POMDP) \cite{gmytrasiewicz2004interactive}.\\
\indent BA-POMDP augments the state space with a state transition count and a state observation count variables as additional hidden states \cite{ross2011bayesian}, \cite{katt2018learning}. It maintains a belief over the augmented state space, resulting in an optimal trade-off between model learning and reward collecting. An implicit assumption of this approach is that the unknown POMDP model is either fixed or varying slower than the model learning process, which is unlikely to be applicable to active perception where an adversary opponent might be learning and adapting too.


An alternative approach is to find a robust policy, which does not assume a fixed model. Robust POMDP assumes the true transition and observation probability belongs to a bounded uncertainty set, and optimizes the policy for the worst case \cite{osogami2015robust}. 

CC-POMDP finds an optimal deterministic policy that satisfies chance constraints. This formulation results in a better trade-off between robustness and nominal performance than Robust POMDP. 
While this formulation shows promising results in challenging risk-sensitive applications\cite{santana2016rao}, we argue that the class of uncertainty considered in the CC-POMDP formulation might be too restrictive for our application. 
However, a better framework that utilizes a wide class of adversary model to find a stochastic optimal policy should exhibit better robustness against an adversary.

I-POMDP extends the POMDP framework by augmenting the hidden space of each agent with a type attribute. The type attribute of an agent includes its preference and belief. The agents reason about the types of the other agents, resulting in nested beliefs. The issue with this approach is that the type space is so large that inference becomes computationally intractable \cite{gmytrasiewicz2004interactive}. In practice, a finite set of possible models for each agent is assumed, which makes it suffer from the difficulty of model incompleteness.

\subsection{Game theoretic frameworks}
The POMDP framework simplifies the active perception problem into a single-agent planning problem. An alternative multi-agent view of this problem is a Bayesian game, where the autonomous agent is unsure about the identity of the opponent. In such a problem, Bayesian Nash Equilibrium strategies are the optimal solutions. Regret Minimization (RM) \cite{zinkevich2008regret} and Fictitious Play (FP) \cite{heinrich2015fictitious} are two algorithmic frameworks for finding Nash Equilibriums. Recent extensions of these two frameworks with deep neural networks achieved success in imperfect information zero-sum games \cite{jin2017regret}, \cite{heinrich2016deep}.
One significant advantage of this framework over POMDP is that no agent model is explicitly required. Instead, the opponent behavior is implicitly specified by the joint payoff (reward), assuming rationality. 
The Nash Equilibrium policy is a robust solution in the sense that it is the best response to a perfect opponent.  
In reality, however, the opponent could be of bounded rationality and reasoning capability, which is non-trivial to model in this game theoretic framework. Moreover, we often have a strong prior over the probable opponent behaviors, while the game theoretic approach completely neglects this knowledge. We expect that algorithms exploiting a reasonable opponent model can outperform the Nash Equilibrium in both the nominal and mildly off-nominal cases. Moreover, convergence of RM and FP in general-sum two-player games is not established, making it difficult to apply these techniques to our active perception problem, since the opponent is not necessarily adversarial, but could be indifferent (e.g., neutral). 
 
\begin{figure}[t]
\centering
\includegraphics[width=0.75\columnwidth]{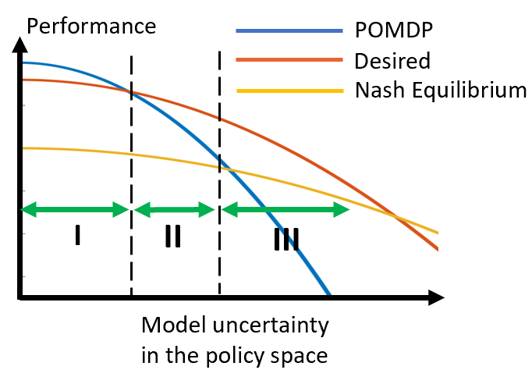} 
\caption{Anticipated correlation between model uncertainty and performances of different active perception algorithms (some evidence shown in Fig. \ref{uncertainty_vs_performance}). While POMDP is optimal given an accurate model, it might be sensitive to model uncertainty. Nash Equilibrium should be robust to model uncertainty, but it might sacrifice too much nominal performance when a good model is available.}
\vspace{-1.5em}
\label{illustration_uncertainty}
\end{figure}

Based on the above discussions, we anticipate a correlation between model uncertainty and the performance of different classes of algorithms, as illustrated in Fig. \ref{illustration_uncertainty}. A desired solution should be near optimal given a good model, and degrades gracefully with respect to increasing model uncertainty.

\subsection{Deep reinforcement learning}
Deep reinforcement learning has led to several recent breakthroughs to solving difficult problems in both MDP \cite{mnih2013playing}, \cite{silver2017mastering} and POMDP \cite{hausknecht2015deep}, \cite{karkus2017qmdp} domains, single and multi-agent \cite{lowe2017multi}, \cite{omidshafiei2017deep}, \cite{peng2017multiagent} domains. The difficulty of deep reinforcement learning in multi-agent domain stems from the non-stationarity of the perceived environment dynamics due to the learning process of other agents. This non-stationarity destablizes value function based methods and causes high variance for policy based methods \cite{busoniu2010multi}. A lot of recent works focus on developing learning algorithms that converge in this multi-agent setting, \cite{lowe2017multi}, \cite{foerster2017stabilising}, \cite{marinescu2014decentralised}, while few works have been done on agent modeling \cite{he2016opponent}. We argue that it is crucial to model the opponent in our active perception problem, because otherwise it becomes challenging, if not impossible, to define exploring behavior. Another benefit of maintaining a model is that the action observation history can be compactly summarized into a belief state, which retains the Markovian property even in partially observable settings. This Markovian property is crucial to the convergence of many reinforcement learning algorithms. 
\section{Approach}
In this section, we describe our algorithm to address the perceived challenges, i.e., active perception robust to un-modeled adversarial strategies. We first formalize the problem description and list the assumptions we make. We then give a general description of our algorithm followed by the implementation detail.\\
\indent We model the active perception problem as a planning problem, defined by the tuple $\langle S,A^a,A^o,T,O,R,b_0,\gamma \rangle$, where $S=\langle S^o,S^p \rangle$ is the state of the world, consisting of the set of observable states $S^o$ and the set of partially observable states $S^p$; $A^a$ is the set of actions of the autonomous agent; $A^o$ is the set of actions of the opponent; we further assume that regardless of the intention, the opponent has the same set of observable actions. Otherwise, an intention is easily identifiable once an action that is uniquely corresponding to that type of intention is observed. $T:S \times A^a \times A^o \rightarrow \Delta_S $ is the transition probability, where $\Delta_{\bullet}$ denotes the space of probability distribution over the space $\bullet$. $O: S \times A^a \rightarrow \Delta_{A^o}$ is the observation probability; $R: S \times A^a \times A^o \rightarrow  \mathbb{R}$ is the reward function; $b_0$ is the prior probability of the opponent being an adversary; and $\gamma$ is the discount factor.\\
\indent We make the following modeling assumptions: 

\begin{enumerate}
\item The opponent is either a neutral or an adversary with hostile intents.
\item A neutral opponent is self-interested, whose behavior can be modeled by a reactive policy $\pi^{\text{neu}}(a^o_t|s_t)$.
\item A hostile opponent is primarily goal-directed, which is defined by a known MDP.
\item A hostile opponent is of bounded rationality, implying that it might not be able to always take the optimal action; moreover, it is likely to behave deceptively in order to achieve its goal.
\end{enumerate}

We also assume that a reasonably accurate neutral behavior model is available. We then generate a parametric set of hostile models with two parameters representing the level of rationality and the level of deception, respectively. We use a feed-forward neural network (NN) to represent the policy of the autonomous agent. This NN takes a binary belief state as its input, which is obtained from Bayesian filtering the hidden intention based on an average model.  The output is a stochastic policy. The reward function is composed of a belief dependent reward, which encourages exploring behavior, and a state dependent reward, which ensures safety. In order to minimize the exploitability, we apply the soft-Q learning \cite{haarnoja2017reinforcement} algorithm that learns a maximum entropy policy.

We present the detail on agent modeling, belief space reward and policy learning in the following subsections.
\subsection{Opponent modeling}
We use a binary variable $\lambda \in \{0,1\}$ to denote whether the opponent is a neutral or an adversary with hostile intents. Depending on $\lambda$, the opponent is expected to exhibit different behaviors, which is fully described by an opponent policy $\pi^o(a^o_t|s_t)$. This model is restrictive since the action probability only depend on the current state. Nonetheless, we use this model only for policy learning, and use a general history dependent opponent policy for the evaluation of the learned autonomous agent policy. Another implicit assumption of this model is that the opponent has full observability over the states. This assumption could be released by modeling the opponent as a POMDP agent.

\noindent \textbf{Neutral model}: If the opponent is a neutral, i.e. $\lambda=0$, we assume a simple reactive policy $\pi^{\text{neu}}(a^o_t|s_t)$ is available to model the opponent:
\begin{equation}
\pi^o(a^o_t|s_t,\lambda=0)=\pi^{\text{neu}}(a^o_t|s_t).
\label{eq1}
\end{equation}
\textbf{Adversary model}: We use the following equation to model an adversarial agent's policy $\pi^o$:
\begin{equation}
\begin{aligned}
\pi^o(a^o_t|s_t,\lambda=1;\alpha,\beta)=&\textnormal{argmin}_{\pi \in \Delta} \{\mathbb{KL}(\pi|\pi^{\text{MDP}}_{\alpha})\\
&+\beta \mathbb{KL}(\pi|\pi^o(\cdot|s_t,\lambda=0)) \},
\end{aligned}
\label{eq2}
\end{equation}
\begin{equation}
\pi^{\text{MDP}}_{\alpha}(a_t^o|s_t,\lambda=1)=e^{\alpha Q(s_t,a_t^o)}/Z(s_t),
\label{eq3}
\end{equation}
where $\mathbb{KL}(\cdot|\cdot)$ denotes the Kullback\textendash Leibler divergence between two distributions, The goal-achieving policy $\pi^{\text{MDP}}_{\alpha}$ is associated with the optimal Q function $Q(s_t,a_t^o)$, of a goal-achieving adversary MDP defined later. The temperature parameter $\alpha$ in (\ref{eq3}) represents the level of rationality of the adversary. 
 The other parameter $\beta$ indicates the level of deception. 
$Z(s_t)$ is the partition function that normalizes $\pi^{\text{MDP}}_{\alpha}$. 

We assume that the goal-directed behavior can be modeled by a known adversary MDP $\langle S^o,A^o,T,R^o,\gamma^o \rangle$, where $S^o=S\times A^a$ is the state space of the active perception problem augmented by the action space of the autonomous agent. This implies that if the autonomous agent takes different actions, the opponent will be in different MDP states, even though the world space $S$ is the same, allowing the opponent taking different responses. $A^o$, the action space of the adversary MDP, is the same as that of the active perception problem, and so is the transition probability $T$. The reward function $R^o$ specifies the reward for the adversary MDP, which is different from that of the active perception. $\gamma^o$ is the discount factor, which could be different from $\gamma$.
The interpretation of (\ref{eq2}) and (\ref{eq3}) is: the adversary policy is a balance between the goal-achieving actions corresponding to the adversary MDP (first term in (\ref{eq2})) and the deceptive actions by imitating the neutral policy (second term in (\ref{eq2})). By varying the two hyper-parameters $\alpha$ and $\beta$, we obtain a set of policies describing a variety of adversary behaviors, which is expected to make the optimized active perception policy more robust. Therefore, we assume a uniform hyper-prior over these two hyper-parameter:
\begin{equation}
p(\alpha,\beta)=\begin{cases} 1 &\mbox{if } 1\leq \alpha \leq 2, 1\leq \beta \leq 2; \\
0 & \mbox{otherwise}. \end{cases}
\end{equation}

\subsection{Belief space reward}
We maintain a belief $b_t(\lambda)$ over the hidden variable by Bayesian filtering. This requires both models for the neutral and the adversary. The neutral model is given by Eq. \ref{eq1}. As the adversary model includes two continuous parameters, inference over the joint space of $\lambda, \alpha, \beta$ is expensive and might not be helpful because this handcraft model may not match the real adversary behavior. We use an average model by marginalizing out the hyper-parameters
\begin{equation}
\bar{\pi}^o(a_t^o|s_t,\lambda=1)=\iint \pi^o(a^o_t|s_t,\lambda=1;\alpha,\beta)p(\alpha,\beta)d\alpha d\beta.
\label{eq4}
\end{equation}
With (\ref{eq1}) and (\ref{eq4}), the Bayesian update rule for the belief is well defined. We define a hybrid belief-state dependent reward to balance exploration and safety
\begin{equation}
\begin{aligned}
r(b_t,s_t,a^a_t)&=-H(b_t)+r(s_t,a^a_t)\\
&=b\log b+(1-b)\log(1-b)+r(s_t,a^a_t),
\end{aligned}
\label{eq6}
\end{equation}
where we use the shorthand $b$ to denote $b_t(\lambda=1)$, the belief that the opponent is an adversary; and $r(s_t,a^a_t)$ is the state dependent reward.
This reward (\ref{eq6}) balances exploration behavior and safety. The negative entropy reward $-H(b_t)$ can be interpreted as maximizing the expected logarithm of true positive rate (TPR) and true negative rate (TNR). The state-dependent reward $r(s_t,a^a_t)$ 
is used to ensure safety. For instance, some actions could be dangerous to the neutral, which are discouraged by a large negative reward.
\subsection{Policy learning}
We use soft-Q learning \cite{haarnoja2017reinforcement} to learn a stochastic belief space policy. 
The soft-Q learning objective is to maximize the expected reward regularized by the entropy of the policy,
\begin{equation}
\sum_t \mathbb{E}_{b_t,s_t,a_t\sim \rho_\pi} \gamma^t[r(b_t,s_t,a_t)+\sigma H(\pi(\cdot|b_t,s_t))].
\end{equation}
\indent The parameter $\sigma$ controls the `softness' of the policy. The nice interpretation of this objective function is maximizing accumulative reward while behaving as uncertain as possible, which is a desired property against an adversary.\\
\indent This maximum entropy problem is solved using soft-Q iteration. For discrete action space, the fixed point iteration:
\begin{equation}
Q_{\text{soft}}(b_t,s_t,a_t)\leftarrow r_t+\gamma \mathbb{E}_{b_{t+1},s_{t+1}\sim p_s} [V_{\text{soft}}(b_{t+1},s_{t+1})],
\label{eq8}
\end{equation}
\begin{equation}
V_{\text{soft}}(b_t,s_t) \leftarrow \sigma \log \sum_{a \in A} \exp(\frac{1}{\sigma}Q_{\text{soft}}(b_t,s_t,a)),
\end{equation}
converges to the optimal soft value functions $Q^*_{\text{soft}}$ and $V^*_{\text{soft}}$ \cite{haarnoja2017reinforcement}, and the optimal policy can be obtained from:
\begin{equation}
\pi^*_{\text{MaxEnt}}(a_t|b_t,s_t)=\exp(\frac{1}{\sigma}(Q^*_{\text{soft}}(b_t,s_t,a_t)-V^*_{\text{soft}}(b_t,s_t))).
\end{equation}

\subsection{Implementation detail}
In order to stabilize the training of the soft value functions, a separate target value network is used, whose parameter is an exponential moving average of that of the value network, with average coefficient $\tau=0.01$. During the training, the value at the right hand side of (\ref{eq8}) is replaced by the value of the target value function.

We use two feedforward neural networks to parametrize the soft Q function and the soft value function. Each neural network has three fully connected hidden layers with 64, 128, and 64 hidden units, respectively, followed by the Relu nonlinear activation function. We use L1 loss and Adam stochastic optimizer with learning rate $lr=5 \times 10^{-4}$ for the value function training. We use a batch size of 50, and an experience replay buffer of size $10^6$, with $10^5$ training epochs. We tested different entropy parameter $\sigma$, and selected $\sigma=0.25$. This value leads to both stable training and decent performance.
The pipeline of our algorithm is summarized by the pseudo code in Algorithm \ref{pseudocode}.
\begin{algorithm}[t]
\caption{Soft-Q learning for active perception}
\begin{algorithmic}
\STATE $B$=[]   \mbox{ } \#initialize the experience replay buffer
\STATE $SQNN.\text{initialize}()$   \mbox{ } \#initialize the soft Q networks
\FOR{$n=1;n \leq N;n=n+1$}
\STATE $\lambda \sim \text{Bernoulli}(b_0)$
\IF{$\lambda=0$}
\STATE $\pi^o\leftarrow \pi^{\text{neu}}$   \mbox{ } \#neutral policy
\ELSE
\STATE $\alpha,\beta \sim p(\alpha, \beta)$\mbox{ } \#adversary hyper-parameter
\STATE $\pi^o\leftarrow$RHS of (\ref{eq2})   \mbox{ } \#adversary policy
\ENDIF

\STATE $B.\text{append}(SQNN.\text{simulate}(\pi^o))$
\STATE $SQNN.\text{train}(B.\text{sample}())$
\ENDFOR
\end{algorithmic}
\label{pseudocode}
\end{algorithm}
 
\section{Case study: threat discrimination at a checkpoint}
\subsection{Problem description}
We evaluate our algorithm via a simple threat discrimination scenario at a checkpoint. In this scenario, an autonomous agent wants to identify if an oncoming opponent is a neutral or an adversary.

\noindent \textbf{States:} The states consist of the fully observable physical state: the distance of the opponent from the checkpoint $d_t$, and the binary hidden state of the opponent, $\lambda$, indicating neutral or adversary.

\noindent \textbf{Actions and observations:} At each time instance, 
the autonomous agent is allowed to take one action from three possible actions: (1) Send a \textbf{hand} signal, (2) Use a \textbf{loudspeaker}, (3) Use a \textbf{flare} bang. The opponent has two possible reactions at each time instance: (1) \textbf{Stay} at the same place (2) Continue \textbf{proceeding} towards to checkpoint.

\noindent \textbf{State transition:} None of the three actions of the autonomous agent has direct effects on the opponent state, while the response of the opponent to these actions are different in the probabilistic sense. If the opponent takes the first action (stay), then its distance from the checkpoint does not change; otherwise, if the opponent takes the second action (proceed), the distance decreases by a unit distance:
\begin{equation}
d_{t+1}=
\begin{cases} d_{t} &a^o_t=\textbf{stay}; \\
d_{t}-1 & a^o_t=\textbf{proceed}. \end{cases}
\end{equation}

\noindent \textbf{State dependent reward:} The state dependent reward for the autonomous agent ($r(s,a^a)$ in (\ref{eq6})) is shown in Table \ref{t2}. Any actions taken upon a neutral is penalized. Aggressive actions (such as \textbf{flare} bang) are penalized more heavily than conservative actions (such as \textbf{hand} signal).

\noindent \textbf{Initial and terminal conditions:} Initially, the opponent is 12 unit distance away from the checkpoint, $d_0=12$. The autonomous agent and the opponent take turns making their actions for 10 rounds, with the autonomous agent taking its action first. This interaction terminates at the 10-th round, i.e., $T=10$.

\noindent \textbf{Opponent agent model:}\\
\indent \textbf{Neutral:} A neutral responds reactively to the actions taken by the autonomous agent, according to the probability shown in Table \ref{t1}.

\indent \textbf{Adversary:} The primary goal of an adversarial opponent is to get close to the checkpoint as quick as possible to conduct a malicious attack. We use the following dense reward for the adversary, which increases as the distance from the checkpoint decreases:
\begin{equation}
r^{\text{MDP}}=1.1^{d_0-d}-1.
\label{eq12}
\end{equation}
We choose $\gamma^o=\gamma=0.95$. The adversary policy is determined by (\ref{eq2}), (\ref{eq3}), and (\ref{eq12}) together.
\subsection{Baseline}
We compare our algorithm with a planning based baseline algorithm, CC-POMDP.
In this framework, the observation probabilities of the POMDP model are assumed to be drawn from a probability distribution. The optimal policy is computed such that the corresponding value function is guaranteed to be higher than a maximized threshold with a certain (high) probability. Formally, the CC-POMDP optimal value function can be found by the iteration
\begin{equation}
V_{n+1}(b)=\max_{a^a \in A^a} [r(b,s,a^a)+\tilde{V}(a^a)],
\end{equation}
where $\tilde{V}(a^a)$ is obtained from the chance constraint optimization problem
\begin{eqnarray}
&& \hspace*{1in}\max_{a^a \in A^a} \tilde{V}, \notag \\
&& \hspace*{-.25in}\text{s.t.~} \mathbb{P}_{\pi^o\sim p_{\pi}}[\gamma\sum_{s\in S} b(s) \sum_{a^o\in A^o} \pi^o(a^o|s,a^a) V_n(b'_{b,a^a,a^o})>\tilde{V}] \notag\\ &&\hspace*{1.5in}>1-\epsilon,
\end{eqnarray}
where $n$ is the iteration index. $\pi^o$ is the modeled observation probability. $p_{\pi}$ is the probability distribution over the observation probability. $b'_{b,a^a,a^o}$ is the updated belief from the belief state $b$, after taking the action $a^a$ and observing the reaction $a^o$. $0\leq \epsilon \leq 1$ is a confidence bound, which typically takes a small value. We choose $\epsilon=0.05$, as this results in the best performance.

\begin{table}[t] 
	\caption{Observation probability of a neutral's reaction} 
	\centering      
	\begin{tabular}{c c c}  
		\hline\hline                        
		Action/Reaction &Stay &Proceed\\ [0.3ex] 
		\hline                    
		Hand & 0.60 & 0.40   \\
		Loudspeaker & 0.75 & 0.25   \\   
		Flare & 0.90 & 0.10   \\   
		\hline     
	\end{tabular}
	\label{t1}  
	\vspace*{.1in}
	\caption{State dependent reward for  active perception agent} 
	\centering      
	\begin{tabular}{c c c}  
		\hline\hline                        
		Action/State &Neutral &Adversary\\ [0.3ex] 
		\hline                    
		Hand & -0.1 & 0   \\
		Loudspeaker & -0.3 & 0   \\   
		Flare & -0.7 & 0   \\   
		\hline     
	\end{tabular}
	\label{t2}  
	\vspace{-1em}
\end{table}

In order to solve the chance constraint optimization, the probability distribution $p_{\pi}$, over the observation probability parameters $\pi^o$ needs to be specified. We model these probability parameters as uncorrelated Gaussian random variables with the nominal probability parameters determined by Eq. \ref{eq1} (with Table \ref{t1}), \ref{eq2}, \ref{eq3}, \ref{eq12} as their mean values. We choose the variance of these Gaussian variables to be $var(\pi^o)=0.001$. 
\subsection{Evaluation criterion}
\begin{figure*}[t]
    \centering
    \begin{subfigure}[b]{0.3\textwidth}
        \centering
        \includegraphics[height=1.5in]{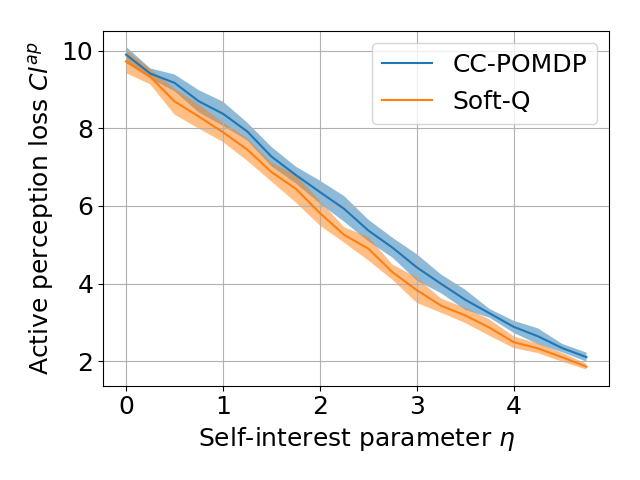}
    \vspace*{-1em}
        \caption{Accumulated active perception loss $Cl^{\text{ap}}$ against \textbf{learning} adversary}
        \label{cl}
    \end{subfigure}%
    ~ 
    \begin{subfigure}[b]{0.3\textwidth}
        \centering
        \includegraphics[height=1.5in]{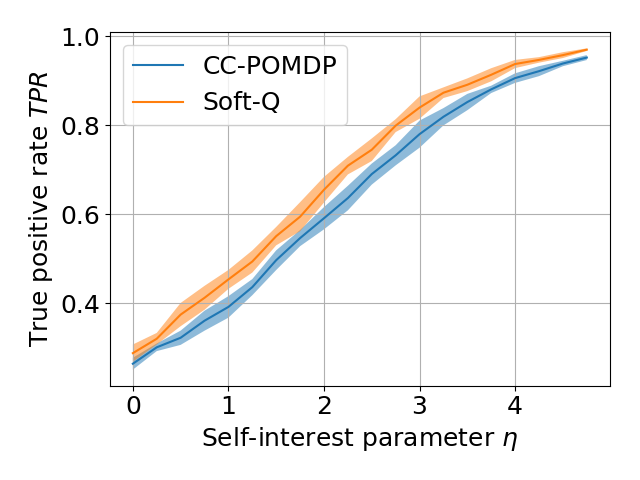}
    \vspace*{-1em}
        \caption{True positive rate (TPR) against \textbf{learning} adversary}
        \label{tpr}
    \end{subfigure}
    ~
    \begin{subfigure}[b]{0.3\textwidth}
        \includegraphics[height=1.5in]{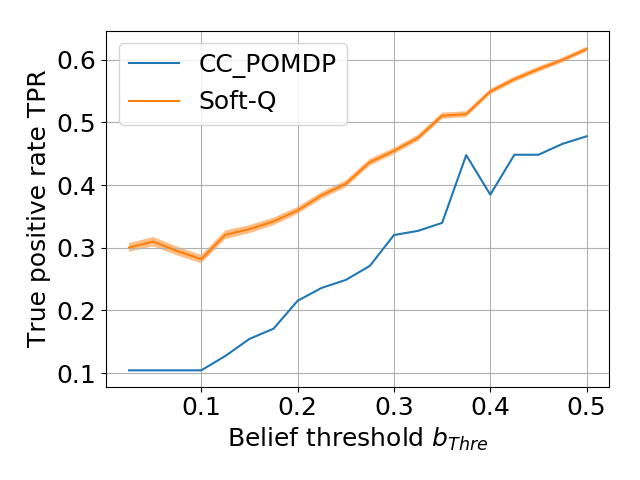}
    \vspace*{-1em}
        \caption{True positive rate (TPR) against \textbf{deceptive} adversary}
        \label{tpr_dece}
    \end{subfigure}
    \vspace*{-.5em}
    \caption{Comparisons of two performance criteria between our proposed algorithm and CC-POMDP, against both a learning adversary ((a) and (b)) parametrized by the self-interest parameter $\eta$ and a deceptive adversary (in (c)) parametrized by $b_{\text{Thre}}$: a small $\eta$ indicates high degree of adversary, and a small $b_{\text{Thre}}$ indicates high degree of deception; Low $Cl^{\text{ap}}$ and high TPR are desired.}
    \vspace{-1.5em}
    \label{performance}
\end{figure*}

In order to evaluate the robustness of the active perception policy against un-modeled adversarial behaviors, we test the performance on two adversary models that have not been presented in the training phase:

\indent \textbf{Learning adversary:} In the first testing adversary model, we apply the soft-Q learning algorithm to train an adversary policy with the following adversary reward
\begin{equation}
r^{\text{adv}}=-r(b,s,a^a)+\eta r^{\text{MDP}},
\end{equation}
where the self-interest parameter $\eta \in [0,+\infty)$ controls the degree of self-interest, and its inverse $\frac{1}{\eta}$ can be interpreted as the degree of adversary. 
We choose different values of $\eta$ in the range $[0,5]$, with step size $\Delta \eta=0.25$. 
We favor the adversary by rendering him the access to the internal belief state of the autonomous agent. The adversary policy network takes this belief state, the immediate action taken by the autonomous agent and the observable distance from the checkpoint, outputting a probability distribution over its two available actions. We define the following two criteria to evaluate the performance of the active perception algorithms
\begin{enumerate}
	\item \textbf{Accumulated active perception loss:}\\ $Cl^{\text{ap}}=-\mathbb{E}[\sum^T_{t=0} \gamma^t r_{t}]$
	\item \textbf{True positive rate (TPR):} \\ $\text{TPR}=\mathbb{E}[b_{t=T}|\lambda=1]$
\end{enumerate}
%
%

\textbf{Deceptive adversary:} In the second testing adversary model, we handcraft a deceptive strategy. A deceptive adversary uses the following policy when the autonomous agent takes actions \textbf{hand} or \textbf{loudspeaker}:
\begin{equation}
a^o=\begin{cases} \textbf{stop} &b_t>b_{\text{Thre}}; \\
\textbf{proceed} & b_t \leq b_{\text{Thre}}, \end{cases}
\end{equation}
while only when the autonomous agent takes action \textbf{flare}, the adversary takes action \textbf{proceed}. $b_{\text{Thre}}$ is a free parameter that controls how deceptive the adversary is. We choose $b_{\text{Thre}}$ in the range $[0,0.5]$, with step size $\Delta b_{\text{Thre}}=0.025$.

These two adversary models are not present in the learning phase. The active perception agent has to identify the hidden intent while being robust to this model uncertainty, which is challenging. 
\subsection{Results}
\textbf{Against a learning adversary:} We present the performance criteria against a learning adversary in Fig. \ref{cl} and \ref{tpr}. These results are averaged over 10 learning simulations. Our algorithm shows better performance compared with the CC-POMDP baseline over the whole range of the self-interest parameter $\eta$, despite the small difference. This is likely attributed to the ensemble modeling used in our algorithm, which results in better robustness against model uncertainty. Nonetheless, planning against an ensemble model is difficult to formulate in the POMDP framework.

\textbf{Against a deceptive adversary:} We present the TPR against a deceptive adversary in Fig. \ref{tpr_dece}. The deterministic CC-POMDP policy achieves significantly lower TPR than the stochastic soft-Q learning policy does. As the CC-POMDP policy deterministically selects action \textbf{hand} when the belief is lower than some threshold, selects \textbf{loudspeaker} otherwise, and never selects \textbf{flare}, the deceptive adversary can maneuver this belief state by exploiting this deterministic structure. In contrast, since the soft-Q learning agent uses a maximum entropy stochastic policy, the adversary's maneuvering strategy becomes less effective. 
\begin{figure}[t]
\centering
\includegraphics[width=0.9\columnwidth]{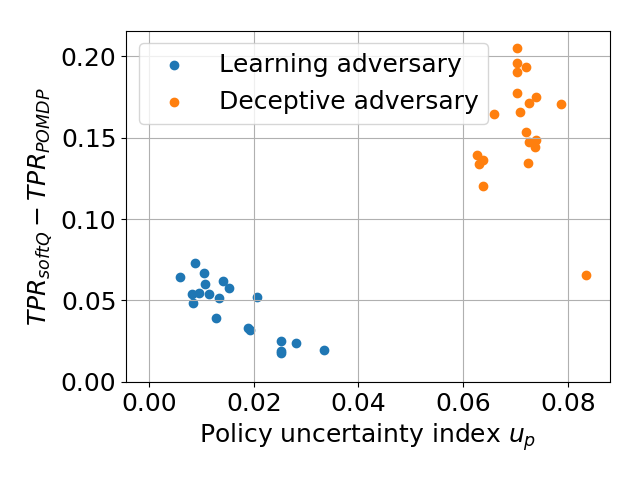} 
    \vspace*{-1.25em}
\caption{Correlation between the adversary model uncertainty and the difference in performance between our algorithm and CC-POMDP: In general, a large policy uncertainty index corresponds to a large difference.}
\label{uncertainty_vs_performance}
\vspace{-0.5em}
\end{figure}

We anticipate that the difference in performance between our algorithm and CC-POMDP is correlated with the adversary model uncertainty, as illustrated in Fig. \ref{illustration_uncertainty}. We provide some evidence to justify this speculation in Fig. \ref{uncertainty_vs_performance}. We defined a policy uncertainty index $u_p$, as one metric to quantify the deviation of the actual adversary policy from the assumed one
\begin{equation}
u_p=\sum_{a^a \in A^a}\sum_{s \in S}\sum_{t=1}^T \frac{\hat{N}_{s,a^a,t}}{\hat{N}\hat{M}}\mathbb{KL}[\hat{\pi}(a^o|s,a^a,t)|\pi(a^o|s,a^a,t)],
\end{equation} 
where $\hat{\pi}$ is an empirical estimation of the adversary policy from data tuples $(a^o,s,a^a,t)$ collected in simulations. $\pi$ is the nominal adversary policy defined in Eq. \ref{eq2}. $\hat{N}_{s,a^a,t}$ is the number of counts that $(\bullet,s,a^a,t)$ appears in the data, where $\bullet$ represents any value for $a^o$. $\hat{N}=\sum_{s,a^a,t}\hat{N}_{s,a^a,t}$ is the total number of data tuples; and $\hat{M}$ is the total number of different data tuples $(\bullet,s,a^a,t)$ appearring in simulations. 

\indent Fig. \ref{uncertainty_vs_performance} shows that, generally speaking, a large model uncertainty corresponds to a large difference in performance. 
This observation explains the small difference between our algorithm and CC-POMDP against the learning adversary shown in Fig. \ref{cl} and \ref{tpr}, and the large difference against the deceptive adversary shown in Fig. \ref{tpr_dece}. It also suggests that our algorithm has the desired property illustrated in Fig. \ref{illustration_uncertainty}, which might be a better trade-off between nominal and off-nominal performance than both POMDP and Nash Equilibrium.  
\section{Conclusion}
In this work, we pose an active perception problem against an opponent with uncertain intent and potential adversarial behaviors. We reviewed related fields of research and pointed out the gap between the existing approaches and the desired solution properties. We then presented a novel solution combining generative adversary modeling, belief space planning, and maximum entropy deep reinforcement learning. Compared with a CC-POMDP baseline, the proposed algorithm is more robust to un-modeled adversarial strategies. One limitation of this work is that we still need to specify an opponent behavior model, which could be non-trivial in complicated applications. We are developing learning algorithms that learn a reasonable opponent model through self-play, to address this limitation. 
\balance
\section*{Acknowledgment}
The authors want to thank Dr.~Kasra Khosoussi for his insightful discussions. This work is supported by ARL
DCIST under Cooperative Agreement Number W911NF-17-2-0181.


\bibliographystyle{IEEEtran}
\bibliography{ursa}
\end{document}